\pdfoutput=1
\documentclass[11pt]{article}

\usepackage[]{emnlp2023}

\usepackage{times}
\usepackage{latexsym}
\usepackage{multirow}
\usepackage[T1]{fontenc}
\usepackage[utf8]{inputenc}

\usepackage{microtype}

\usepackage{inconsolata}
\usepackage{amsmath}
\usepackage{subcaption}
\usepackage{booktabs}
\usepackage{xcolor}
\usepackage{xspace}
 
\newcommand{\alg}{LASCL\xspace}
\newcommand{\scl}{\text{SCL}\xspace}
\newcommand{\algii}{\text{LI}\xspace}
\newcommand{\algic}{\text{LIC}\xspace}
\newcommand{\algisc}{\text{LISC}\xspace}
\newcommand{\algiuc}{\text{LIUC}\xspace}

\usepackage{kky}
\usepackage{cleveref}
\newcommand{\Scref}[1]{\S\ref{#1}}
\usepackage{enumitem}

\title{Learning Label Hierarchy with Supervised Contrastive Learning}

\author{Ruixue Lian \qquad William A. Sethares \qquad Junjie Hu \\
  University of Wisconsin-Madison \\
  \texttt{\{ruixue.lian, sethares, junjie.hu\}@wisc.edu} \\
  }

\begin{document}
\maketitle

\begin{abstract}
Supervised contrastive learning (SCL) frameworks treat each class as independent and thus consider all classes to be equally important. This neglects the common scenario in which label hierarchy exists, where fine-grained classes under the same category show more similarity than very different ones. This paper introduces a family of Label-Aware SCL methods (\alg) that incorporates hierarchical information to SCL by leveraging similarities between classes, resulting in creating a more well-structured and discriminative feature space. This is achieved by first adjusting the distance between instances based on measures of the proximity of their classes with the scaled instance-instance-wise contrastive. An additional instance-center-wise contrastive is introduced to move within-class examples closer to their centers, which are represented by a set of learnable label parameters. The learned label parameters can be directly used as a nearest neighbor classifier without further finetuning. In this way, a better feature representation is generated with improvements of intra-cluster compactness and inter-cluster separation. Experiments on three datasets show that the proposed \alg~works well on text classification of distinguishing a single label among multi-labels, outperforming the baseline supervised approaches. Our code is publicly available.\footnote{\url{https://github.com/rxlian/LA-SCL}}

\end{abstract}
\section{Introduction}
\label{sec:introduction}

% \jh{Create an outline here (check this tutorial: \url{https://cs.stanford.edu/people/widom/paper-writing.html}) }
\begin{figure}[htbp!] 
\centering
    \includegraphics[width=0.9\linewidth]{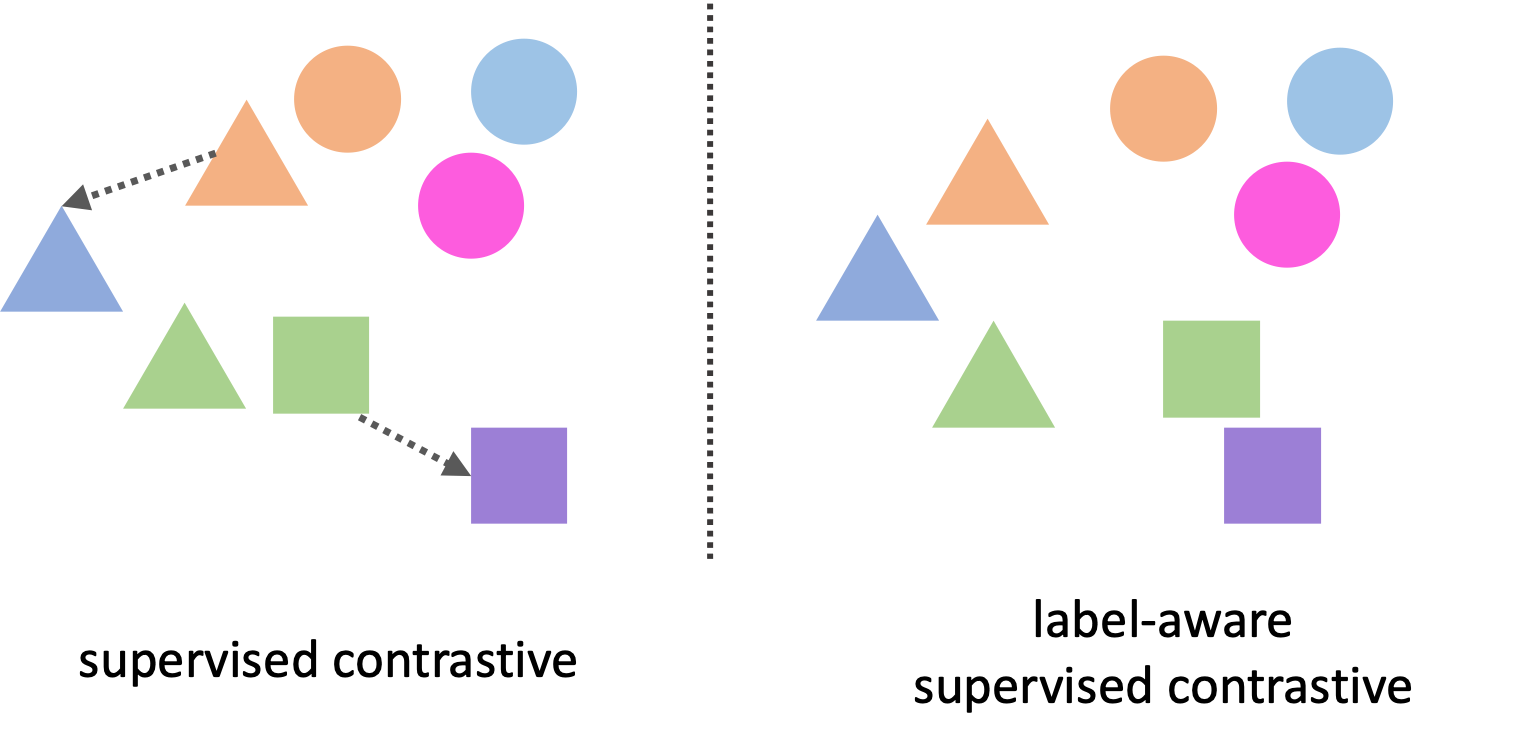}
 \caption{Supervised v.s. label-aware supervised contrastive loss: The supervised contrastive loss (left) contrasts the set of all samples from the same class as positives against the negatives from the remainder of the batch \cite{khosla2020supervised}. The label-aware supervised contrastive loss (right) proposed in our work incorporates label hierarchy by considering class similarities.}
 \label{fig: overview}
\end{figure}
Supervised contrastive learning (SCL) \cite{khosla2020supervised} aims to learn generalized and discriminative feature representations given labeled data. It relies on the construction of positive pairs from the same class and negative pairs from different classes, thereby encouraging similar data points to have similar representations while pushing dissimilar data points apart in the feature space. This method considers each class to be independent and considers all classes to be of equal importance, thus treating the problem without awareness of any relationships among the labels. However, in the real world, it is natural that class labels may relate to each other in complex ways, in particular, they may exist in a hierarchical or tree structure \cite{malkinski2022multi,demszky2020goemotions,murdock2016blockout,verma2012learning,han2018hierarchical}. Within a data hierarchy, different sub-categories under the same branch tend to be more similar than those from different branches, since they will tend to have similar high-level semantics, sentiment, and structure. This similarity should be reflected in the feature representations. 

Hierarchical text classification (HTC) is one way to structure textual data into a tree-like category or label hierarchy, representing a taxonomy of classes \cite{kowsari2017hdltex}. Existing HTC can be divided into global and local approaches. Global approaches treat the problem as a flat classification, while local approaches build classifiers for labels at each level of the hierarchy. \citet{an2022fine} propose FCDC, which aims to transfer information from coarse-grained levels to fine-grained categories and thus adapt models to categories of different granularity. Besides, \citet{wang2022incorporating}~incorporate label hierarchy information extracted from a separate encoder. Some other works leverage additional hierarchical information \cite{lin2023effective,long2023facilitating,suresh2021not}. 

Other than that, \citet{zeng2023learning} augment the classification loss by the Cophenetic Correlation Coefficient (CPCC) \cite{sokal1962comparison} as a standalone regularizer to maximize the correlation between the label tree structure and class-conditioned representations. \citet{li2020prototypical} propose a ProtoNCE loss, a generalized version of the InfoNCE loss \cite{oord2018representation} to learn a representation space by encouraging each instance to become closer to an assigned prototype such as the clustering centroid. In this way, the underlying semantic structure of the data can be encoded.

Based on these studies, the hierarchical structure of the labels suggests that learning methods could be enhanced if the learning mechanism can be made aware of the class taxonomy. We explore several ways of exploiting such hierarchical relationships between classes by proposing to augment the SCL loss function as depicted in Fig. \ref{fig: overview}. Since this incorporates class taxonomy information, we call it label-aware SCL (\alg). This is achieved by first using pairwise class similarities to scale the temperature in the SCL to encourage samples under the same branches to cluster more closely while driving apart samples with different labels under different coarse clusters. In addition, we add instance-center-wise contrastive with learned label representations as the center of the sentence embeddings from the corresponding class. These result in making sub-classes under the same coarse-grained classes closer to each other and generating more discriminative representations by making intra-class samples closer to their centers.

To utilize intrinsic information from label and data hierarchies, we encode the textual label information to be class centers and compute pairwise class Cosine similarities on top of that. This quantifies the proximity between classes and forms the basis for instantiating variations of \alg objectives. Since the dimension of these label representations is the same as the linear classifier, we show that it can be applied directly to downstream classification without further finetuning. To the best of our knowledge, we are the first to work on leveraging the textual hierarchical label and integrating it into the SCL to improve the representations. Our methods can be transferred to various backbone models, and are simple yet effective across different datasets. The only changes we make are in the cost function so that the method can be applied in any situation where labels in a hierarchy exist.

Our contributions are summarized as follows:
\begin{itemize}[leftmargin=10pt,topsep=1pt] \itemsep-0.1em
    \item \alg~integrates label hierarchy information into SCL by leveraging the textual descriptions of the label taxonomy.
    \item Our method learns a structured feature space by making fine-grained categories under the same coarse-grained categories closer to each other.
    \item Our method also encourages more discriminative representations by improving intra-cluster compactness and inter-cluster separation.
    \item The learned label parameters from our method can be used directly as a nearest neighbor classifier without further finetuning.
   
\end{itemize}

% Section 2 describes the problem setup, the goals of the learning process, and a brief overview of the SCL framework. Section 3 describes the algorithmic innovations of this paper, which incorporate label-specific information into the loss function of the learning  process. Section 4 demonstrates the effectiveness of the method in two experiments which have hierarchical label structures. Section 5 provides the main experimental results which show the effectiveness of the label-aware strategies in providing better (more compact) clusters and in separating different clusters from each other in the embedding space. This is done both numerically and visually. Section 5 also presents studies that explore how the various members of the \alg~family compare. Section 6 reviews the literature on the use of label hierarchies in learning methods, and the literature on contrastive learning. Section 7 concludes. 
\section{Background}
\label{sec:prelim}
%\jh{Describe the data setup and the goal of learning. for example, "Given a dataset of xxx, we aim to learn an embedding function that ..."}

\paragraph{Problem Setup} For a supervised classification task, a labeled dataset $\mathcal{D}=\{(x_i, y_i)\}_{i=1}^N$ consists of $N$ examples from a joint distribution $P_{\mathcal{XY}}$, where $\mathcal{X}$ is the input space of all text sentences, $\mathcal{Y}=\{1, ..., C\}$ is the label space, and $C$ is the number of classes. The goal of representation learning is to use $\mathcal{D}$ to learn a feature encoder $f_\theta: \mathcal{X}\to\mathcal{Z}$ that encodes a text sentence to a semantic sentence embedding in a feature space $\mathcal{Z}$. This allows us to measure the pairwise similarity between two text sentences $x_i, x_j$ by a similarity function $\text{sim}(x_i, x_j)$, which first projects $x_i$ and $x_j$ to $\Zcal$, i.e., $\zb_i=f_\theta(x_i)$, and computes a distance between two sentence embeddings in $\Zcal$. Moreover, learning meaningful embeddings facilitates the learning of a classifier $g_\phi: \Zcal \to \Ycal$ that maps learned embeddings to their corresponding labels.

% Given $\mathcal{D}$, we aim to learn an embedding objective function $L$ such that by applying a feature encoder $f_\theta$, a more discriminative feature representation $z=f_\theta(x)$ can be generated. Let $\mathcal{Z}$ be the feature space, which is obtained by $f_\theta: \mathcal{X} \rightarrow \mathcal{Z}$. Let $\text{sim}(z_i, z_j)$ be the similarity measurement between $z_i$, $z_j$.

\paragraph{Supervised Contrastive Learning (SCL)} A major thread of representation learning focuses on supervised contrastive learning~\cite{khosla2020supervised} that encourages embedding proximity among examples in the same class while simultaneously pushing away embeddings from different classes using the loss function in Eq. \eqref{eq:label-supcl}. Specifically, for a given example $(x_i, y_i)$, we denote $\Pcal(y_i)=\{x_j| y_j=y_i, (x_j, y_j)\in \Dcal\}$ as the set of sentences in $\Dcal$ having the same label as $y_i$. Thus, the SCL loss is computed on $\Dcal$ as:

% {\small
% \begin{align} \label{eq:label-supcl} 
%     \nonumber & \ell_\scl(x_i, y_i) = \frac{1}{|\Pcal(i)|}\sum\limits_{j\in P(i)} \log \frac{\exp(\frac{\text{sim}(\zb_i, \zb_j)}{\tau})}{\sum\limits_{k \notin P(i)} \exp(\frac{\text{sim}(\zb_i, \zb_k)}{\tau})}  \\ 
%     &\Lcal_{\tau}(\Dcal; \theta) = \frac{1}{N} \sum_{i=1}^N \ell_{\tau}(x_i, y_i),
% \end{align}
% }

\begin{align} \label{eq:label-supcl} 
    \nonumber & \resizebox{\linewidth}{!}{ $ \ell_\scl(x_i, y_i) = \mathbb{E}_{x_j\sim \Pcal(y_i)} \log \frac{\exp(\frac{\text{sim}(x_i, x_j)}{\tau})}{\sum\limits_{k \notin \Pcal(y_i)} \exp(\frac{\text{sim}(x_i, x_k)}{\tau})} $}\\ 
    % &\Lcal_{\tau}(\Dcal; \theta) = \frac{1}{N} \sum_{i=1}^N \ell_{\tau}(x_i, y_i),
    &\Lcal_{\tau}(\Dcal; \theta) = \mathbb{E}_{(x_i,y_i)\sim \Dcal}~\ell_{\tau}(x_i, y_i),
\end{align}
The fixed hyper-parameter $\tau$ is the temperature that adjusts the embedding similarity of sentence pairs. %The loss function \eqref{eq:label-supcl} forms the baseline loss into which we incorporate the label information in the next section.

% \begin{align} \label{eq:label-supcl}
%     \ell &= -\sum_i \frac{L_\text{SCL}}{|P(i)|} \\
%     L_\text{SCL} &= \sum_{j\in P(i)} \log \frac{\exp(\frac{\text{sim}(z_i, z_j)}{\tau})}{\sum_{k \notin P(i)} \exp(\frac{\text{sim}(z_i, z_k)}{\tau})} \nonumber
% \end{align}
\section{Method}
\label{sec:method}
% In the embedding space, sentences from the same class or semantically similar classes are closer to each other when compared to sentences from completely different classes. In addition, certain classes may lie close to each other while others are relatively far apart. This is especially common when the labels have a hierarchical structure, since subclasses under the same high-level label are likely to be more similar than those from different high-level labels. 

This section describes our proposed label-aware supervised contrastive learning objectives.

\paragraph{Overview:} In the embedding space, we hypothesize that sentences from different fine-grained classes under the same coarse-grained class are closer to each other in comparison to sentences from different high-level categories. Given this intrinsic information provided by the label and data hierarchy, we use the pairwise cosine similarities of a set of learnable parameters representing label features to quantify the proximity between classes, which are used to instantiate variants of label-aware supervised contrastive learning objectives. 
% This label similarity matrix is a set of learnable parameters initialized from the label embedding extracted by a pretrained lauguage model (\Scref{sec:method:label}).

\subsection{Label Hierarchy and Class Similarities}
\label{sec:method:label}
\begin{figure*}[htbp!]
\centering
\begin{subfigure}[b]{0.5\linewidth}
    \centering
    \includegraphics[width=1\linewidth]{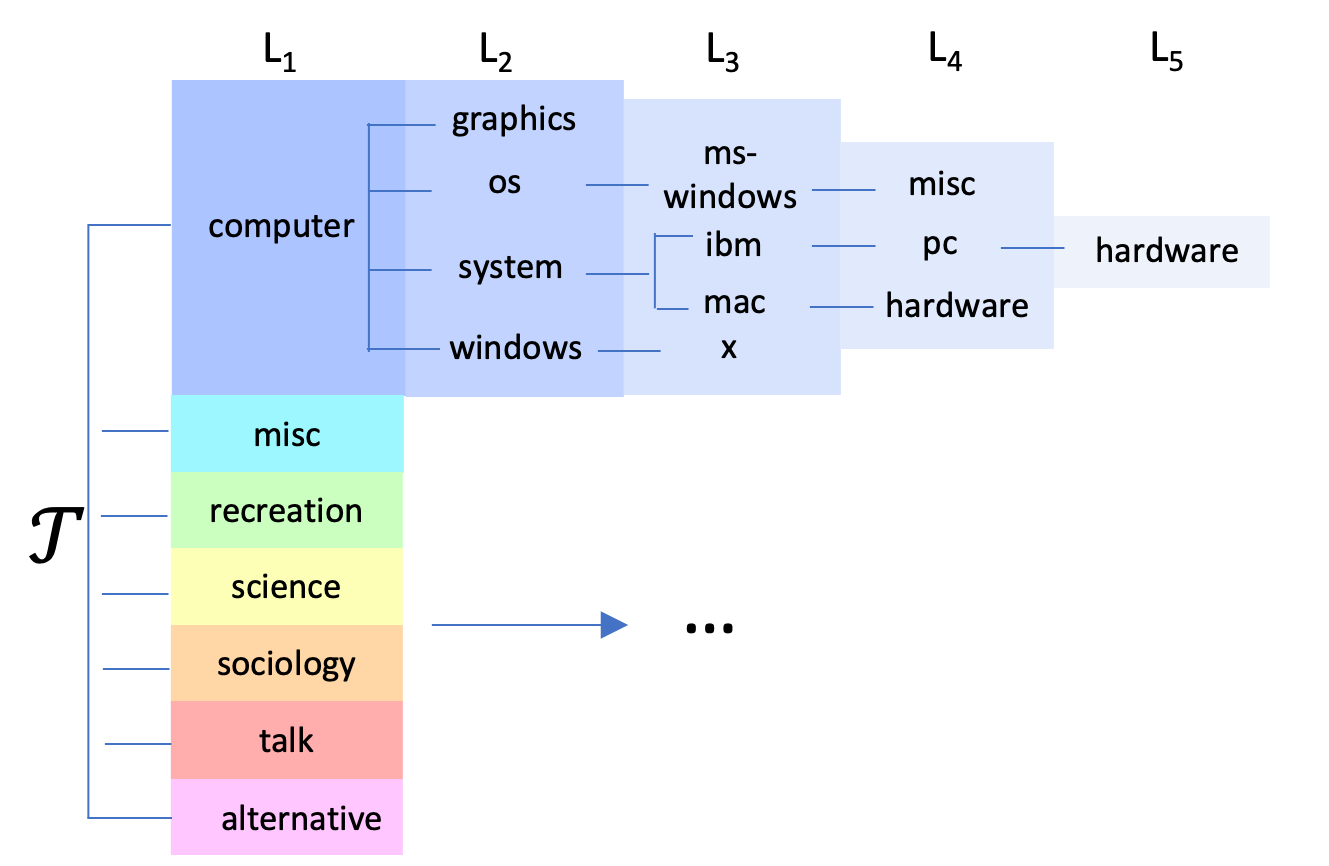}
    \caption{}
    \label{fig: tree}
\end{subfigure}\hspace{-0.5cm}
\begin{subfigure}[b]{0.5\linewidth}
    \centering
    \includegraphics[width=1\linewidth]{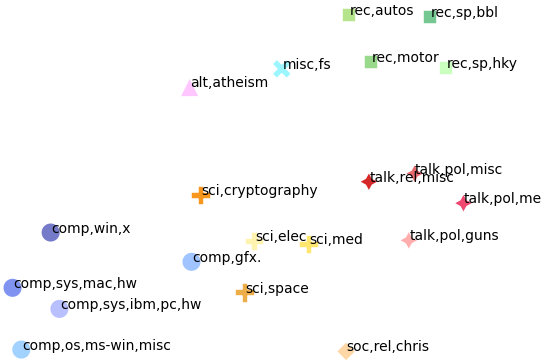}
    \caption{}
    \label{fig: label vis}
 \end{subfigure}
\label{fig: tree and vis}
 \caption{(a) The label hierarchy of the 20News dataset. The root node contains 7 classes, each branch has multiple fine-grained sub-categories. (b) t-SNE visualization of hierarchical label embeddings encoded by BERT-base.}
\end{figure*}

% \begin{figure*}[htbp!]
%   \centering
%   \begin{minipage}[b]{0.22\textwidth}
%     \includegraphics[width=1.25\textwidth]{figures/figs/tree.png}
%     \subcaption{}
%     \label{fig: tree}
%   \end{minipage}\hfill
%   \begin{minipage}[b]{0.22\textwidth}
%     \includegraphics[width=1.25\textwidth]{figures/figs/label_vis.png}
%     \subcaption{}
%     \label{fig: label vis}
%   \end{minipage}
%   \caption{(a) Label hierarchy of the 20NewsGroups dataset. The root node contains 7 classes. Each branch has multiple fine-grained sub-categories. (b) t-SNE visualization of hierarchical label embeddings encoded by BERT-base.}
%   \label{fig: tree and vis}
% \end{figure*}
This section describes the construction of learnable label representations given label hierarchies, which are used to calculate similarities between classes. 

A label hierarchy of a labeled dataset refers to a hierarchical tree that defines an up-down, coarse-to-fine-grained structure with labels being assigned to a corresponding branch. We use label textual descriptions to construct the tree structure. Let $\Tcal$ be a hierarchical tree with $V$ being the set of intermediate and leaf nodes. Each leaf node $v_c$ represents a class label $c\in \Ycal$, and is associated with a set of examples in class $c$, i.e., $\Pcal(c)$, where $\Pcal(c)\cap \Pcal(c')=0$, $\forall c\neq c'$. Each parent node represents a coarse-grained category containing a set of fine-grained children nodes. The leaf nodes can have different depths in $\Tcal$, which refers to the distance between each leaf node $v_c$ and root node $v_0$. Let $L_i$ be the $i$-th layer of $\mathcal{T}$. Figure \ref{fig: tree} shows an example of a tree-structured label hierarchy built from 20News dataset~\cite{lang1995newsweeder}. 
% It has 20 classes in 7 high-level categories. Let $L_i$ be the $i$-th layer of $\mathcal{T}$. Sentences can be first categorized into the 7 major classes: ``Computer, Recreation, Science, Sociology, Talk, Alternative, Misc'' at the $L_1$ level. Moving forward to the subsequent levels, each category can be further split into various fine-grained groups. For example,  samples in ``Computer'' can be splited into four fine-grained groups: ``Graphics, OS, System, Windows.'' %In this way, the classes are partitioned in to a hierarchically. 

Given $\Tcal$, we exploit the hierarchical relationships among the classes by having more informative descriptions. To achieve this, given a leaf node of class $c\in \Ycal$, its ancestor nodes are first collected until reaching the leaf node. These up-down textual classes at different levels are concatenated into a text sequence, which is then filled in by a sentence template. For Figure \ref{fig: tree}, for a leaf node of ``Hardware'' at $L_5$, we collect its ancestors and assign ``Computer, System, IBM, PC, Hardware'' as its label. In this way, the hierarchical information of labels is collected and can be extracted by an encoder. Let $u_c$ be a sentence of class $c\in \Ycal$. A pretrained language encoder $f_\theta$ is used to obtain a label representation denoted as $\ub_c=f_\theta(u_c)$. This set of label representations are made of learnable parameters and will be updated during back-propagation. To stabilize the process, we re-encode the label representations less frequently than the updates of the sentence embeddings, that is, extract label embeddings only after every $n$ iterations. 
%for every several iterations of updates with less frequency instead of encoding the label representations using the latest checkpoint of $f_\theta$ in every iteration.

After encoding label representations for all classes $\Ub=[\ub_1, \dots, \ub_C]$, a pairwise cosine similarity measurement is applied to compute a class similarity matrix $\Wb\in\mathbb{R}^{C\times C}$, where each entry is the similarity score between a label $c$ and another label $c'$, i.e., $w_{cc'}=\text{sim}(u_c, u_{c'})$. $\Wb$ will be further applied to scale the temperature in \Scref{section:label_scale}. Note that this label embedding matrix $\Ub\in\mathbb{R}^{d\times C}$ can be directly used as a nearest-neighbor classifier, where it can be applied to linearly map an input sentence embedding $x_i\in\mathbb{R}^d$ into the label space $\Ycal$. Therefore, $\Ub$ can be applied as a linear head for the downstream classification without further finetuning. 

Figure \ref{fig: label vis} shows the t-SNE \cite{van2008visualizing} visualization of 20 initialized label embeddings of the 20News extracted from their sentence description encoded by a pretrained BERT-base model. Different high-level and lower-level classes are displayed with different markers and colors. Observe that labels from the same coarse-grained classes are clustered closer to each other than to other classes. Given the clustering nature of the labels reflects their hierarchical structure, these class similarities can be utilized as additional information to scale the importance of different classes, which is introduced in the next section. 

\subsection{Scaling with Class Similarities} \label{section:label_scale}
This section describes a way to incorporate the class hierarchy information into supervised contrastive loss by leveraging additional scalings introduced in $\Wb$. The overall idea is to scale the temperature $\tau$ in Eq. \eqref{eq:label-supcl} by $\Wb$, which reflects similarities between classes and is updated every several iterations. Specifically, the negative example pairs in SCL are weighted by the corresponding learned class similarities, performing a scaled instance-to-instance update. The final loss over a dataset $\Dcal$ is the same form as Eq.~\eqref{eq:label-supcl} with the individual loss $\ell_\tau$ replaced by
% \begin{center}
% \resizebox{\linewidth}{!}{$
    \begin{align}
        \resizebox{0.88\linewidth}{!}{ \label{eq:label-sensitive-supcl} 
        $\ell_{sii}(x_i, y_i) = \mathbb{E}_{j\sim \Pcal(y_i)} \log \frac{\exp\left(\frac{\text{sim}(x_i, x_j)}{\tau}\right)}{\sum\limits_{k \notin \Pcal(y_i)} \exp\left(\frac{\text{sim}(x_i, x_k)}{\tau \cdot s_{ik}}\right)}$},
    \end{align}
% \end{center}
where the elements of the matrix $\Wb$ define the pairwise similarity between labels, abbreviated by $s_{ik}=w_{y_i,y_k}$ for a label pair $y_i$ and $y_k$.

In this way, Eq. \eqref{eq:label-sensitive-supcl} scales the similarity between negative pairs based on the similarity between the corresponding classes. Consider two samples $x_i$ and $x_k$ from different classes $y_i$ and $y_k$. The similarity $s_{ik}$ tends to be greater if $y_i$ and $y_k$ have the same parent category. Thus, it applies a higher penalty to the negative pairs when they are from different coarse-grained categories, so the learning update tends to push them further apart. In this way, the label hierarchical information is introduced to assign different penalties, reflecting the similarities and dissimilarities between classes. 
% Similarly, if applies a lower penalty to the negative pairs when they are from the same coarse-grained categories but in different fine-grained classes. 

% Similarly, the temperature in $\ell_{ic}$ can be scaled by the label similarity $s_{ik}$, and thus we can construct a scaled instance-center-wise contrastive loss term $\ell_{sic}$ in Eq.~\eqref{eq:instance-scaled-centroid}.
% \begin{align} \label{eq:instance-scaled-centroid}
%     \ell_{sic}(x_i, y_i) = \log \frac{\exp(\frac{\text{sim}(x_i, u_i)}{\tau})}{\sum_{k \notin P(i)} \exp(\frac{\text{sim}(x_i, u_k)}{\tau \cdot s_{ik}})}.
% \end{align}

\subsection{Label Representations as Class Centers}

The label representations can also be used as class centers to perform instance-center-wise contrastive learning, as shown in another loss term $\ell_{ic}$.
%as shown in Eq.~\eqref{eq:instance-centroid}. 
\begin{align}\label{eq:instance-centroid} 
    \ell_{ic}(x_i,y_i) = \log \frac{\exp\left(\frac{\text{sim}(x_i, u_{y_i})}{\tau}\right)}{\sum_{k \notin \Pcal(i)} \exp\left(\frac{\text{sim}(x_i, u_{y_k})}{\tau}\right)}.
\end{align}
This loss term $\ell_{ic}$ regards the label sequence $u_c$ constructed for the label $c$ as the center of the sentences from this class. Thus, for each input instance $x_i$, a positive pair is constructed between the instance and its center as $(x_i, u_{y_i})$, and negative pairs are constructed by comparing the instance $x_i$ with other label sequences, $(x_i, u_{y_k}), \forall y_k\neq y_i$. This loss function pulls each sentence closer to its label center and further from other centers, thus making each cluster more compact in the embedding space.

Similarly to Eq.~(\ref{eq:label-sensitive-supcl}), the temperature in $\ell_{ic}$ can be scaled by the class similarity $s_{ik}$, and thus we can construct a scaled instance-center-wise contrastive loss term as follow:
%$\ell_{sic}$ in Eq.~\eqref{eq:instance-scaled-centroid}.
\begin{align} \label{eq:instance-scaled-centroid}
    \ell_{sic}(x_i, y_i) = \log \frac{\exp\left(\frac{\text{sim}(x_i, u_i)}{\tau}\right)}{\sum_{k \notin P(i)} \exp\left(\frac{\text{sim}(x_i, u_k)}{\tau \cdot s_{ik}}\right)}.
\end{align}

\subsection{Label-Aware SCL Variants}

\label{sec:method:supcl}
Based on the aforementioned loss functions, we propose four label-aware SCL (\alg) variants and compare their performance in~\Scref{sec:analysis}.
\paragraph{Label-aware Instance-to-Instance (LI)} The first variant is shown in Eq.~\eqref{eq:label-sensitive-supcl}, which modifies the original SCL by scaling the temperature by the label similarity.

\paragraph{Label-aware Instance-to-Unweighted-Center (LIUC)} The second variant augments the original SCL by adding an unweighted instance-center-wise contrastive loss.
% The second variant, which we call the {\em instance-to-unweighted-center} denoted $\ell_{\algiuc}$, which is denoted as
\begin{align} \label{eq:instance-to-unweighted-center}
    &\ell_{\algiuc} = \ell_{\scl} + \ell_{ic}
\end{align}
\paragraph{Label-aware Instance-to-Center (LIC)} The third variant augments our first variant by adding an unweighted instance-center-wise contrastive loss.
% We call the third variation the {\em instance-to-center}:
\begin{align} \label{eq:instance-to-unweighted-center}
    &\ell_{\algic} = \ell_{sii} + \ell_{ic}
\end{align}
\paragraph{Label-aware Instance-to-Scaled-Center (LISC)} The final one augments our first variant by adding a weighted instance-center-wise contrastive loss.
% We call the fourth variation the {\em instance-to-scaled-center} denoted $\ell_{\text{LISC}}$ as follows:
\begin{align} \label{eq:instance-to-unweighted-center}
    &\ell_{\algisc} = \ell_{sii} + \ell_{sic}
\end{align}

\section{Experimental Settings}
\label{sec:setting}

\paragraph{Datasests}
\begin{table}[htbp!]
    \centering
     \resizebox{1\linewidth}{!}{
    \begin{tabular}{c c c c}
    \toprule
         \multirow{2}{*}{Dataset} & train/val/test & train/val/test & classes  \\
         &  (original) (K) &  (LP) (K) & ($|\text{L}_1|$/$|\text{L}_n|$) \\
         \midrule
        20News & 10/1/7 & 2/2/7 &7/20\\
        WOS & 38/4/4 & 1/1/4 & 7/134 \\
        DBPedia &238/2/60 & 12/12/60 & 9/70 \\
        % 20News & 10,183/1,131/7,532 & 2,263/2,263/7,532  &7/20\\
        % WOB & 38057/4299/4699 & 1410/1368/4699 & 7/134 \\
        % DBPedia &238,533/2,409/60,794 &12,048/12,048/60,794  &9/70 \\
    \bottomrule
    \end{tabular}}
    \caption{Dataset statistics. $|\text{L}_1|$ and $|\text{L}_n|$ are number of coarse-grained and fine-grained classes, respectively.}
    \label{tab:data_stat}
\end{table}
20NewsGroups\footnote{\url{http://qwone.com/~jason/20Newsgroups/}} (news classification)~\cite{lang1995newsweeder}, WOS (paper classification) \cite{kowsari2017hdltex}, DBPedia (topic classification)\cite{auer2007dbpedia}, and their originally provided label structures and textual labels are used in our experiments. Each leaf node label of 20News has different depth, while each leaf node lable of WOS and DBPedia have the same depth 2. Dataset statistics is shown in Table \ref{tab:data_stat}. For linear-probe (LP) experiments, we randomly select samples with balanced distribution.

\paragraph{Sentence Templates} We use the following templates to fill in the label string for each dataset, which is further encoded by a BERT model.
% We use the following sentence templates to fill in labels. 
\begin{itemize}[leftmargin=10pt,topsep=1pt] \itemsep-0.1em
    \item 20News: ``It contains \{$\text{label}_i$\} news.''
    \item WOS: ``It contains article in domain of \{$\text{label}_i$\}.''
    \item DBPedia: ``It contains \{$\text{label}_i[L_{2}]$\} under \{$\text{label}_i[L_1]$\} category.''
\end{itemize} 

\paragraph{Implementation Details}
We use \textit{bert-base-uncased} provided in huggingface's packages \cite{wolf2019huggingface} as our backbone models. The averaged word embeddings of the last layer are used as sentence representations. We used learning rate 1e-5 with linear scheduler and weight decay 0.1. The model is trained with 20 epochs and validated every 256 steps. To avoid overfitting, the best checkpoints were selected with an early stop and patience of 5 according to evaluation metrics. For LP, we use a learning rate of 5e-3 with a weight decay of 0.01. The classifier was trained with $10$ epochs and validated after each epoch. The best checkpoint was selected according to validation accuracy. The batch size and max sequence length are 32 and 128, respectively, across all the experiments. The temperature $\tau$ is 0.3. During training, we re-encode the label embeddings every 500 steps. Cosine similarity was used over all experiments.

\paragraph{Evaluation Metrics}
We report: (1) classification accuracy on the leaf node called \textbf{nodeAcc} (2) classification accuracy on the parent node of the leaf, which is called \textbf{midAcc}, (3) classification accuracy on the root node, which is the highest level of each branch and is called \textbf{rootAcc}.

% \paragraph{}

\section{Results and Analysis}
\label{sec:analysis}

To demonstrate the effect of the amount of labeled data to~\alg, we perform experiments with both the few-shot setup and full dataset in \Scref{sec:fewshot} and \Scref{sec:full setup}. In \Scref{sec:res:vis}, we visually show how the proposed methods generate a more well-structured and discriminative embedding space by visualizations. We discuss how the size of the hierarchy plays a role by constructing a bottom-up label hierarchy with different depths in \Scref{sec:sensitivity}.

The experimental results are reported with linear probes (LP) and with direct testing (DT). For LP, a randomly initialized linear layer was trained on a small number of labeled samples with the encoder frozen. We denote DT as directly applying the learned label parameters as the classifier (\Scref{sec:method:supcl}).

\subsection{Few-Shot Cases} \label{sec:fewshot}
% \begin*{figure}[htbp!]
%   \centering
%   \begin{minipage}[b]{0.33\textwidth}
%     \includegraphics[width=1.1\textwidth]{figures/figs/20news_fewshot.png}
%     \subcaption{}
%     \label{fig:news_fewshot}
%   \end{minipage}\hfill
%   \begin{minipage}[b]{0.22\textwidth}
%     \includegraphics[width=1.1\textwidth]{figures/figs/dbp_fewshot.png}
%     \subcaption{}
%     \label{fig:dbp_fewshot}
%   \end{minipage}
%   \begin{minipage}[b]{0.22\textwidth}
%     \includegraphics[width=1.1\textwidth]{figures/figs/wos.png}
%     \subcaption{}
%     \label{fig:wos_fewshot}
%   \end{minipage}
%   \caption{NodeAcc of DT experiment with k-shot learning (a) 20News (b) DBPedia (c) WOS}
%   \label{fig:fewshot}
% \end*{figure}

\begin{figure*}
    \centering
    \begin{subfigure}{0.325\textwidth}
        \includegraphics[width=\linewidth]{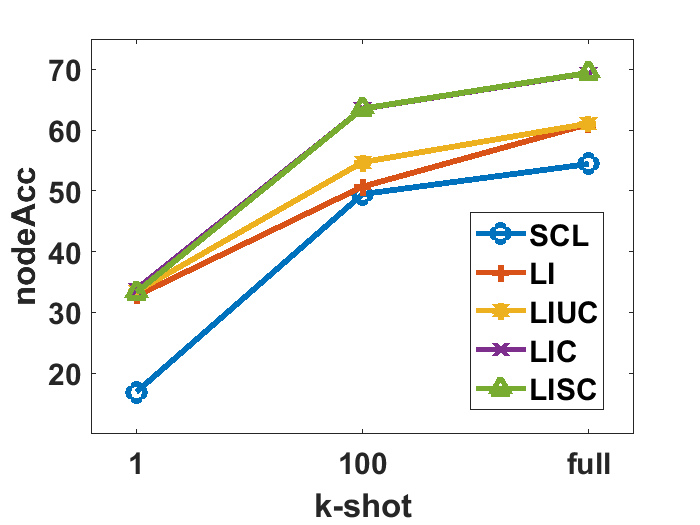}
          \caption{20News}
          \label{fig:news_fewshot}
      \end{subfigure}
  % \centering
    \begin{subfigure}{0.325\textwidth}
    \includegraphics[width=\linewidth]{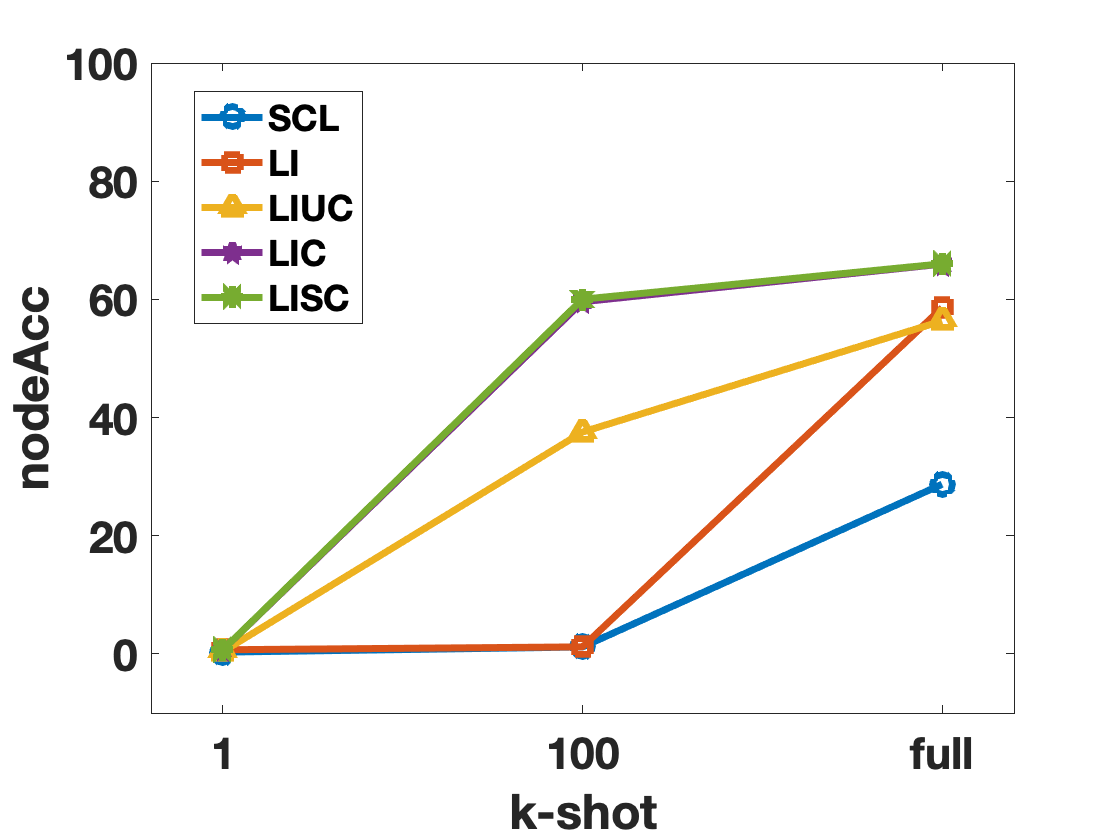}
          \caption{WOS}
          \label{fig:wos_fewshot}
    \end{subfigure}
  \begin{subfigure}{0.325\textwidth}
        \includegraphics[width=\linewidth]{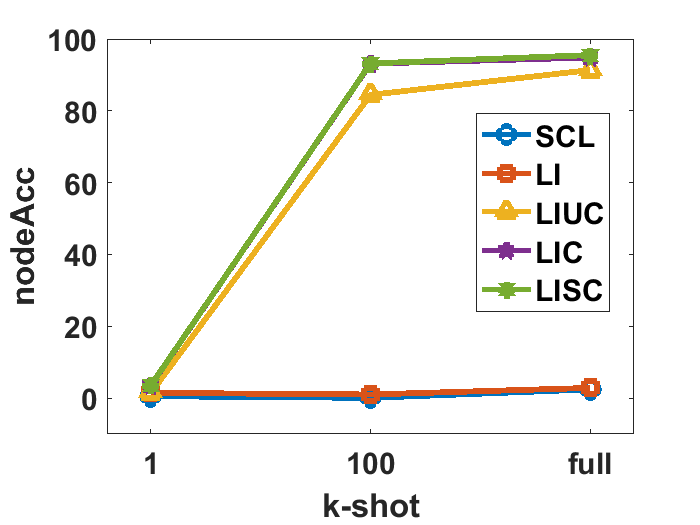}
        \caption{DBPedia}
    \label{fig:dbp_fewshot}
  \end{subfigure}
  
      % \vspace{-0.35cm}
    \caption{Directly testing (DT) the k-shot prediction performance (measured by NodeAcc) on three datasets.}
  \label{fig:fewshot}
    % \vspace{-0.4cm}
\end{figure*}
\begin{table*}[thb!]
    \centering
    \begin{tabular}{cc c c c c c c}
    \toprule
    \multirow{2}{*}{Dataset} & \multirow{2}{*}{Objective} & \multicolumn{3}{c}{direct test} & \multicolumn{3}{c}{linear probe} \\ 
    \cmidrule{3-8}  
        & & nodeAcc & midAcc & rootAcc & nodeAcc & midAcc & rootAcc \\
        \midrule
        \multirow{5}{*}{20News} & SCL & 54.44&61.74&69.41& 65.64&72.54&78.98 \\
        & LI & 61.01&67.19&73.09 & 67.59&74.04&79.82 \\
        & LIUC & 61.09&69.62&79.17 & 66.42&73.66&79.67 \\
        & LIC & 69.40&75.64&81.05 & 68.32 &75.21&80.87 \\
        & LISC & \textbf{69.45}&\textbf{75.90}&\textbf{81.08} & \textbf{68.47}& \textbf{75.33}&\textbf{81.07} \\
        \midrule
        % \multirow{6}{*}{WOS} & FCDC \cite{an2022fine} & 63.64 \\
       \multirow{5}{*}{WOS} & SCL & 28.71 & -- & 46.50 & 54.03 & -- &70.06 \\
        & LI & 58.57& -- & 70.91 & 62.14 & -- & 74.97  \\
        & LIUC & 56.35 & -- & 71.89 & 58.32 & -- & 72.89 \\
        & LIC & 65.97 &--&78.46  & 73.17 &--&83.12 \\
        & LISC & \textbf{66.02}&--&\textbf{78.47} & \textbf{73.56}&--&\textbf{83.13} \\
        \midrule
        \multirow{5}{*}{DBPedia} & SCL & \phantom{0}2.42 & -- & 38.26 & 96.00 & -- & 96.79 \\
        & LI & \phantom{0}2.84 & -- & 31.25 & 96.14&--&96.80 \\
        & LIUC & 91.34 & -- & 94.65 & 96.00&--&96.79\\
        & LIC & 94.85 & -- & 96.30 & 96.52&--&97.25 \\
        & LISC & \textbf{95.52} & -- & \textbf{97.06} & \textbf{96.71}&--&\textbf{97.35} \\
        \bottomrule
    \end{tabular}
    \caption{Classification accuracy (\%) in terms of the leaf, mid-layer, and root nodes with models trained on SCL, LI, LIUC, LIC, and LISC on 20News, WOS, and DBPedia datasets.}
    % \caption{main result \jh{use toprule/midrule/bottomrule instead of hline, avoid vertical lines if possible}}
    \label{tab:main}
\end{table*}

\textbf{\alg works well on few-shot cases.} We first conduct k-shot experiments with k=1 and k=100. To be specific, we take 1 and 100 sentences from each class to construct the training set. The validation and test sets remain the same as the original. NodeAcc on direct testing experiments are shown in Figure \ref{fig:fewshot}, and the accuracies are summarized in Table \ref{tab:few-shot} in the Appendix. 

We can observe improvements under few-shot cases by applying \alg across three datasets, while there are some differences in terms of hierarchical label granularities reflected by the datasets. \algii is effective when there exists a more comprehensive label hierarchical information as shown in Fig. \ref{fig:news_fewshot}, where 20News has a deeper hierarchy of fine-grained labels compared to DBPedia and WOS (Fig. \ref{fig:dbp_fewshot} and \ref{fig:wos_fewshot}) which have only two layers for each label. It indicates that a more comprehensive hierarchy that captures the intricate relationships between classes would be more beneficial.

Besides, \algic, \algiuc, and \algisc, which incorporate additional contrastive objectives between instances and centers, achieve notable performance and largely close the gap, especially between full dataset and 100-shot on DBPedia and WOS datasets. It effectively utilizes the label information even if the hierarchical structure is shallow. With 100-shot, the computation cost is decreased by reducing the training set size to 1\% while maintaining decent performance compared to with full dataset.

\subsection{Full Dataset} \label{sec:full setup}
\textbf{\alg outperforms \scl in full-data setting.} Table \ref{tab:main} shows the results on the full dataset with our proposed four \alg objectives, which outperform SCL in terms of the accuracy on the leaf node, mid-layer, and root level metrics for both DT and LP experiments. In most cases, LP enhances the performance compared to DT, while maintaining a comparable performance across different objectives. The performance gain introduced by \algic and \algisc is substantial enough to narrow the performance gap between DT and LP. In particular, DT performs better than LP on 20News, indicating the creation of effective label representations.

% \begin{figure*}[htbp!]% 
% \centering
% % \begin{subfigure}[t]{0.4\linewidth}
% %     \centering
% %     \includegraphics[width=1\linewidth]{figures/figs/bert_base.png}
% %     \caption{}
% %     \label{fig: bert-base}
% % \end{subfigure}\hfil% equal to outside spacing
% \begin{subfigure}[t]{0.4\linewidth}
%     \centering
%     \includegraphics[width=0.8\linewidth]{figures/figs/supcl.png}
%     \caption[b]{}
%     \label{fig: supcl}
%  \end{subfigure}   \hspace{-0.4cm}
% \begin{subfigure}[t]{0.4\linewidth}
%     \centering
%     \includegraphics[width=0.8\linewidth]{figures/figs/scaled_ii.png}
%     \caption{}
%     \label{fig: scaled_ii}
% \end{subfigure}  \hspace{-0.4cm}
% \begin{subfigure}[t]{0.4\linewidth}
%     \centering
%     \includegraphics[width=0.8\linewidth]{figures/figs/scaledi_scaledc.png}
%     \caption[b]{}
%     \label{fig: scaledi_scaledc}
%  \end{subfigure}
%  \caption{t-SNE visualization of 20NewsGroups when using different loss functions. (a) bert-base, (b) SCL, (c) $\text{LA}_{ii}$, (d) $\text{LA}_{isc}$. Labels are marked by ``x''.}
%  \label{fig: scatter}
% \end{figure*}

\begin{figure*}[htbp!]
  \centering
  \begin{subfigure}[b]{0.35\textwidth}
    \centering
    \includegraphics[width=\textwidth]{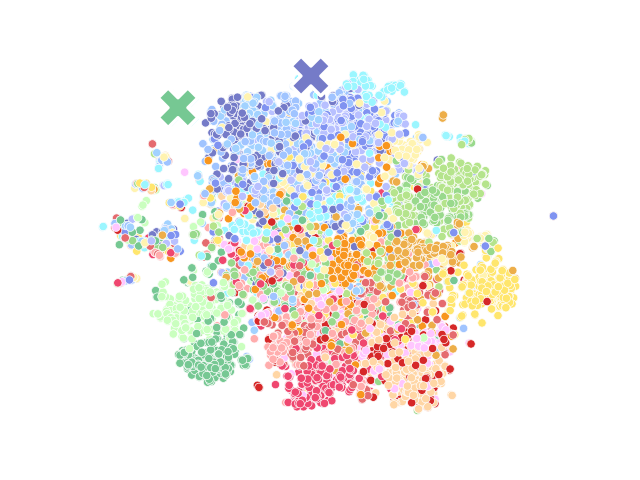}
    \caption{bert-base}
    \label{fig: bert-base}
  \end{subfigure}
  \hspace{-1cm}
  \begin{subfigure}[b]{0.35\textwidth}
    \centering
    \includegraphics[width=\textwidth]{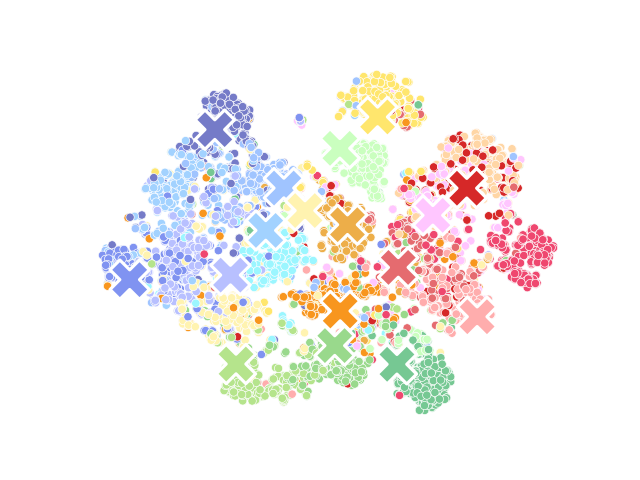}
    \caption{SCL}
    \label{fig: supcl}
  \end{subfigure}
  \hspace{-1cm}
  \begin{subfigure}[b]{0.35\textwidth}
    \centering
    \includegraphics[width=\textwidth]{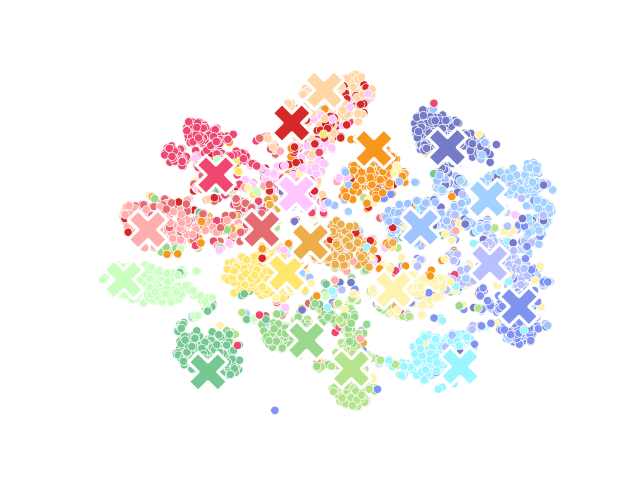}
    \caption{LISC}
    \label{fig: scaledi_scaledc}
  \end{subfigure}
 \caption{t-SNE visualization on 20News dataset (keep the original distribution) with (a) bert-base, (b) SCL, (c) \algisc. Label representations are marked by appropriately colored ``$\times$''. }
 \label{fig: scatter}
\end{figure*}

Among the four proposed variants, the additional scaling introduced by the class similarities contribute to the performance gains, especially when dealing with fine-grained hierarchies.  The improvement is clearest using the nodeAcc test comparing SCL and \algii where the accuracy is increased by effectively penalizing the distance between classes. Moreover, compared to SCL, the additional instance-center-wise contrastive loss introduced by \algiuc also induces performance gains, especially on rootAcc of coarse-grained categories. It leads to clearer decision boundaries between coarse-grained categories, and moves within-class instances closer to their centers. \algic contributes to a further improvement on both nodeAcc and rootAcc by combining the aforementioned two advantages. In contrast, compared to \algic, \algisc provides only a marginal improvement by weighing the class centers because it only introduces small adjustments in the feature space. Further detailed comparison of these methods is presented in \Scref{sec:res:vis}.

\subsection{Visualization} \label{sec:res:vis}
\textbf{\algisc generates a more well-structured and discriminative representation space.} Figure~\ref{fig: scatter} shows a scatter plot of sentence and label embeddings, marked by dots and colored ``$\times$'' respectively, and colored by classes. The distribution of the sampled examples in the figure is the same as the original dataset. Figures \ref{fig: bert-base} - \ref{fig: scaledi_scaledc} show the representations extracted from bert-base, SCL, and \algisc, respectively. We find that \algisc generates a better representation than SCL by bringing clusters belonging to the same high-level classes closer to each other while simultaneously separating clusters of different classes. For instance, consider samples under the coarse-grained class ``recreation'' depicted in green. Initially, in Figure \ref{fig: supcl}, these sub-categories are widely dispersed. While in Figure \ref{fig: scaledi_scaledc}, the four sub-categories of ``recreation'' have become grouped closer to each other. This shows that penalizing the weights between classes with the class similarity matrix effectively guides the model to bring related sub-categories together. This can be interpreted to be a consequence of the ability of \algisc to exploit dependencies among the classes, instead of considering each class independently as SCL does. In addition, the \algisc also mitigates issues when there exist common themes where the corresponding label embeddings overlap one another.

\begin{table}[htbp]
    \centering
    \begin{tabular}{c c c}
    \toprule
        Method & IntraCluster $\downarrow$ & InterCluster $\uparrow$  \\
        \midrule
        SCL & 14.59 & 22.96 \\
        \algii & 14.32 & 23.66 \\
        \algiuc & 14.04 & 23.21 \\
        \algic & 13.62 & 24.31 \\
       \algisc & \textbf{13.52} & \textbf{24.48} \\
        \bottomrule
    \end{tabular}
    \caption{Averaged inter- and intra-cluster $L_2$ distances on 20News, which measure the compactness and separation of clusters, respectively.}
    \label{tab:compact}
\end{table}
To quantitatively demonstrate the effectiveness of these methods, we calculate the average pairwise $L_2$ intra- and inter-cluster distances on 20News to measure the compactness of each cluster and distance between clusters as shown in Table \ref{tab:compact}. Smaller intra-cluster distance implies a more compact cluster. Meanwhile, the clusters are well-separated with a larger inter-cluster distance. Comparing SCL and \algiuc, we can see that the additional instance-center-wise contrastive particularly improves cluster compactness by moving within-class examples closer to their centers. Comparing SCL to \algii shows that the inter-cluster distance increases by applying class similarity to scale the temperature, leading to a more discriminative embedding space. \algisc achieves the best performance among all variations by combining the aforementioned advantages. As a result, \algisc facilitates clearer decision boundaries and improves the representation and organization in the embedding space.

\subsection{Sensitivity to Different Label Hierarchies} \label{sec:sensitivity}
\textbf{Deeper hierarchical structures work better.} To demonstrate the effect of hierarchy size, we assess how each leaf node label performs under different hierarchical structures. By manipulating the layers of the labels, we simulate different levels of granularity. To achieve this, we construct different label hierarchies with bottom-up levels ranging from 1-5 on 20News. The performance is always measured on the leaf nodes to make a fair comparison. 

% \begin{figure}[!tbp]
%   \centering
%     \includegraphics[width=0.35\textwidth]{figures/figs/hie1.png}
%   \caption{NodeAcc measured on different bottom-up label hierarchies ranging from 1-5.}
%   \label{fig:hie1}
% \end{figure}

\begin{figure}[htbp!]
  \centering
  \begin{subfigure}{0.236\textwidth}
        \includegraphics[width=\linewidth]{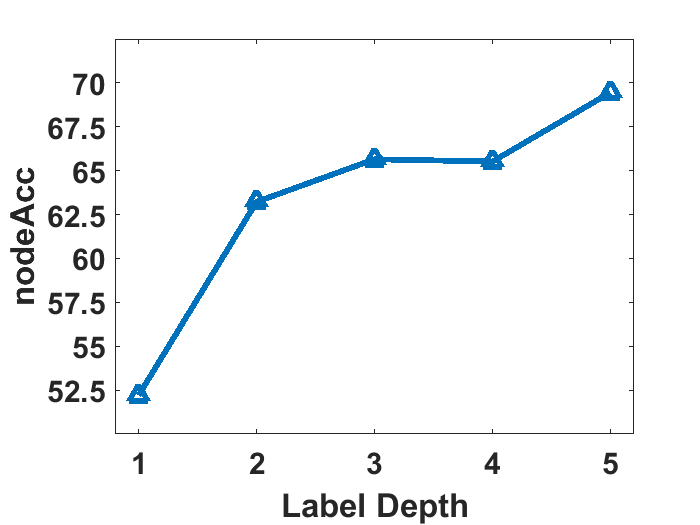}
    \subcaption{}
    \label{fig:hie1}
  \end{subfigure} \hfill
  \begin{subfigure}{0.236\textwidth}
        \includegraphics[width=\linewidth]{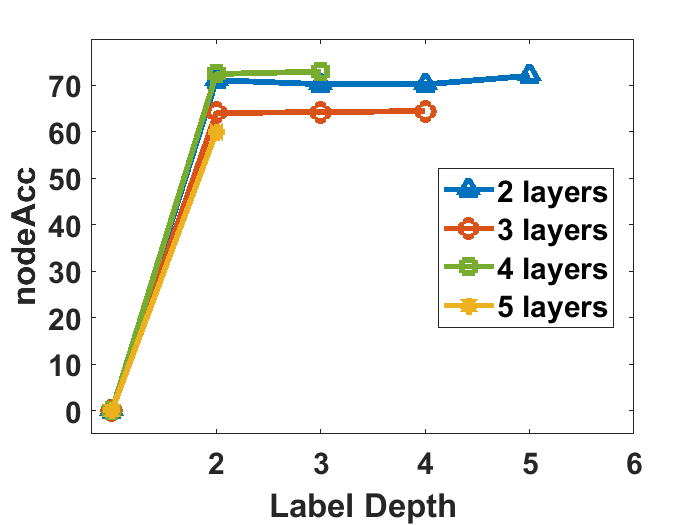}
    \subcaption{}
    \label{fig:hie3}
  \end{subfigure}
  \caption{Measure the sensitivity to different hierarchies on 20News in (a) nodeAcc with different bottom-up label hierarchies ranging from 1-5. (b) nodeAcc on labels grouped by different hierarchies.}

\end{figure}

We observe that the overall performance changes in response to different levels of label granularity, as shown in Figure \ref{fig:hie1}. A similar observation can be found in Figure~\ref{fig:hie3}, which groups the performance based on the hierarchy of leaf nodes with depths ranging from 2-5. From Figure~\ref{fig:hie3}, we notice that the model makes more precise predictions with more specific label information as the hierarchical depth increases. Besides, the proposed methods can also be applied to flat labels when the label depth is 1 given that we can leverage the label description as long as we have that prior knowledge. Thus, the model can better distinguish between closely related classes when provided with more detailed comprehensive labels.

\section{Related Work}
\label{sec:related}
\paragraph{Learning Label Hierarchy}
Hierarchical text classification is a task involving assigning samples to specific labels (most commonly fine-grained levels) arranged in a structured hierarchy, which is typically represented as a tree or directed acyclic graph, where each node corresponds to a label \cite{pulijala2004hierarchical}. Recent studies have suggested integrating the label structure into text features by encoding them with a label encoder. For instance, \citet{chen2020hyperbolic} embed the word and label hierarchies jointly in the hyperbolic space. \citet{zhou2020hierarchy} propose a hierarchy-aware global model to extract the label structural information. \citet{zhang2022hcn} design a label-based attention module to extract information hierarchically from the labels on different levels. \citet{wang2022incorporating} propose a network to embed label hierarchy to text encoder with contrastive learning. \citet{chen2021hierarchy} propose a matching network to match labels and text at different abstraction levels. Other than these studies on network structure, \citet{ge2018deep} propose a hierarchical triplet loss, which is useful for finding hard negatives by hierarchically merging sibling branches. Recent work by \cite{zhang2022use} introduces a hierarchy-preserving loss, applying a hierarchical penalty to contrastive loss with the preservation of a hierarchical relationship between labels on images by using images under the same branch as positive pairs. Our \alg, in contrast, exploits a small number of known labels and their hierarchical structure to improve the learning process. It differs from these works in constructing penalties from the hierarchical structure and exploiting it in the contrastive loss.

\paragraph{Contrastive Learning}
Self-supervised contrastive learning is a representation learning approach that maximizes agreement between augmented views of the same instance and pushes different instances far apart. Works on text data \cite{rethmeier2023primer} constructing various augmentations on text level \cite{wu2020clear,xie2020unsupervised,wei2019eda,giorgi2020declutr}, embedding level \cite{wei2019eda, guo2019augmenting,sun2020mixup,uddin2020saliencymix}, and via language models \cite{meng2021coco,guo2019augmenting,chuang2022diffcse}, etc. SCL effectively learns meaningful representations and improves classification performance by combining supervised and contrastive learning advantages. It was initially introduced in SimCLR \cite{chen2020simple}. Other following works introduce novel insights to improve the representation learning such as MoCo \cite{he2020momentum}, BYOL \cite{grill2020bootstrap}, and SwAV \cite{caron2020unsupervised}. SCL has also been applied to NLP tasks such as sentence classification \cite{chi2022conditional}, relation extraction \cite{li2022hiclre,chen2021cil} and text similarity \cite{zhang2021pairwise,gao-etal-2021-simcse}, where it has shown promising results in learning effective representations for text \cite{sedghamiz2021supcl,khosla2020supervised, chen2022contrastnet}.

\paragraph{Multi-label classification} Multi-label text classification is to assign a subset of labels to a given text \cite{patel2022modeling,giunchiglia2020coherent}. It acknowledges that a document can belong to more than one category simultaneously, and is especially useful when dealing with complex and diverse content that may cover multiple topics or themes. The modeling dependencies amongst labels in this work only consider assigning a single category to each sequence, and our future study is to extend this method to multi-label classification.
\section{Conclusion}
\label{sec:conclusion}
In this work, we propose \alg to include information about the label hierarchy by introducing scaling to the \scl loss to penalize distances between negative example pairs using the class similarities constructed from the learned label feature representations. An additional instance-center-wise contrastive is introduced. These bring instances with similar semantics or belonging to the same high-level categories closer to each other, encourage each instance to become closer to its centers, and the underlying hierarchical structures can be encoded. A better-structured and discriminative feature space is generated by improving the intra-cluster compactness and inter-class separation. The learned labeled parameters can be directly applied as a nearest neighbor classifier without further tuning. Their effectiveness is demonstrated with experiments on three text classification datasets.

\section*{Limitations}
\label{sec:limitations}
Our proposed methods have some limitations, particularly when dealing with highly fine-grained label structures where most of the branches exhibit significant similarities. In this case, it is challenging to distinguish between label embedding similarities. Assigning weights to different classes may not be effective since the similarity scores $w_{cc'}$ are almost identical. This hinders the ability to accurately differentiate between classes and further impacts the performance. Another limitation comes from the common underlying issue of data. Bias can be learned by the model. To mitigate this, debias techniques can be employed to ensure fair and unbiased representation.

% \input{sections/ethical}

\bibliography{anthology,custom}
\bibliographystyle{acl_natbib}
\newpage

\appendix

\section{Appendix}
\label{sec:appendix}
\subsection{LP with Label Embeddings}

In the experiments of Section 5, we randomly initialized the parameters of the classifier. An alternative is to use the pretrained label-representative parameters as the linear head, and then to further train on the labeled dataset used in the linear probe. Results on 20NewsGroups are shown in Table \ref{tab:label-lp}. Comparing their performance to Table \ref{tab:main}. Further tuning the label embedding matrix on labeled samples with cross-entropy loss impairs the performance with \algii and \algiuc. It achieves comparable or slightly better performance in terms of \algisc and \algic.
\begin{table}[htbp]
    \centering
    \begin{tabular}{c|c c c}
    \toprule
       Objective & nodeAcc & midAcc & rootAcc \\
        \midrule
        \algii & 67.26&73.74&78.78 \\
        \algiuc & 64.42&68.08&78.45 \\
        \algic & 68.99&72.90&80.75 \\
        \algisc & 69.15&76.00&81.40 \\
        \bottomrule
    \end{tabular}
    \caption{(\%). LP by using label embeddings as an initialized classifier on 20NewsGroups.}
    \label{tab:label-lp}
\end{table}

\subsection{Sensitivity on Different Label Templates}
\begin{table*}[h!]
    \centering
    \begin{tabular}{c c ccc ccc}
    \toprule
    \multirow{2}{*}{Templates} & \multirow{2}{*}{Objective} & \multicolumn{3}{c}{directly test} & \multicolumn{3}{c}{linear probe} \\ 
    \cline{3-8} 
        & & nodeAcc & midAcc & rootAcc & nodeAcc & midAcc & rootAcc \\
        \midrule
        \multirow{4}{*}{1} & \algii & 61.35 & 64.63 & 76.62 & 58.47 & 65.75 & 74.50 \\
        & \algiuc & 67.66 & 75.31 & 79.93 & 58.30 & 65.53 & 74 .44\\
        & \algic & 63.39 & 71.92 & 80.35 & 57.79 & 65.52 & 74.08 \\
        & \algisc & 67.34 & 75.66 & 79.43 & 57.78 & 65.44 & 74.16 \\
        \midrule
        \multirow{4}{*}{2} & \algii & 66.62 & 73.43 & 78.98 & 94.62& -- &93.69\\
        & \algiuc & 67.49 & 74.79 & 79.65 &94.66& -- &95.66\\
        & \algic &65.45 & 73.88 & 80.02 &94.25& -- &95.35\\
        & \algisc & 68.35 & 75.11 & 79.61 &94.25& -- &95.35 \\
        \midrule
        \multirow{4}{*}{3} & \algii & 65.43 & 72.29 & 78.52 & 66.88 & 73.62 & 79.13 \\
        & \algiuc & 67.69 & 74.88 & 80.24 &94.66& -- &95.66\\
        & \algic & 64.70 & 73.25 & 80.20 & 65.69 & 73.39 & 79.02\\
        & \algisc & 67.90 & 75.00 & 79.49&94.25& -- &95.35 \\
       
        \bottomrule
        
    \end{tabular}
    \caption{Results with different label templates on 20News.}
    \label{tab:few-shot}
\end{table*}
We explore the sensitivity of different label templates on 20NewsGroups as an example. Other than the template used in section \Scref{sec:setting}, we also use the following templates
\begin{enumerate}
    \item This sentence delivers $\{\text{label}_i\}$ news under the category of $\{\text{label}_i[L_1]\}$
    \item Description of $\{\text{label}_i\}$ by generating a sentence from ChatGPT, the prompt given to ChatGPT is ``Please generate a sentence to describe $\{\text{label}_i\}$ news.''
    \item $\{\text{label}_i\}$: description of $\{\text{label}_i\}$
\end{enumerate}

In 2nd template, we use ChatGPT to generate a sentence description for each label. For instance, the description of ``recreation,sport,hockey'' is ``In the latest recreation and sport news, hockey enthusiasts are buzzing with excitement as teams gear up for an intense season filled with thrilling matches and adrenaline-pumping action on the ice.''

% \subsection{Individual Label Performance over Different Label Hierarchies}
% \input{figures/hie_appendix}

% This appendix assesses how each parent node label performs with different hierarchical contexts. As shown in Figure \ref{fig:hie2}, we measure the performance on 6 parent nodes with models trained with different top-down numbers of label hierarchies. Labels such as ``misc'' and ``recreation'' exhibit consistently high performance across all hierarchies. Among them, ``misc'' only has two levels with one branch and is easy to classify. 

% On the other hand, there may be labels that show a notable improvement in performance as the hierarchy becomes more granular such as ``alternative,'' ``computer,'' and ``talk.'' Among them, ``alternative'' only has two levels with one branch. The categories ``computer'' and ``talk'' benefit from a more fine-grained hierarchy as the performance increases when the hierarchical level becomes deeper. These labels benefit from the additional layers in the hierarchy, as the increased specificity enables the model to capture their unique characteristics better.
 
% The category ``science'' has four leaf nodes corresponding to four sub-categories, all on the second layer in the tree. That is the reason why the performance jumps across layer 1 to 2 and remains stable afterward.

\subsection{Comprehensive Few-Shot Cases Results}
\begin{table*}[h!]
\small
    \centering
    \begin{tabular}{c c ccc ccc}
    \toprule
    \multirow{2}{*}{Dataset} & \multirow{2}{*}{Objective} & \multicolumn{3}{c}{directly test} & \multicolumn{3}{c}{linear probe} \\ 
    \cline{3-8} 
        & & nodeAcc & midAcc & rootAcc & nodeAcc & midAcc & rootAcc \\
        \midrule
         \multicolumn{8}{c}{1-shot} \\
        \midrule
        \multirow{5}{*}{20News} & SCL & 16.89 & 22.81 & 42.06 & 58.68 & 66.60 & 74.97 \\
        & \algii & 32.71 & 41.20 & 56.03 & 58.47 & 65.75 & 74.50 \\
        & \algiuc & 33.43 & 41.66 & 57.32 & 58.30 & 65.53 & 74 .44\\
        & \algic & 33.82 & 42.11 & 57.47 & 57.79 & 65.52 & 74.08 \\
        & \algisc & 33.30 & 40.96 & 56.47 & 57.78 & 65.44 & 74.16 \\
        \midrule
        \multirow{5}{*}{WOS} & SCL &0.32& -- &12.22 &34.39& -- &52.05 \\
        & \algii & 0.70& -- &14.43 & 49.94& -- &66.08\\
        & \algiuc & 0.41& -- &13.30 &49.33& -- &65.18\\
        & \algic &0.71& -- &14.07 &50.20& -- &66.16\\
        & \algisc & 0.70& -- &14.47&50.69& -- &66.23 \\
        \midrule
        \multirow{5}{*}{DBPedia} & SCL &0.52& -- &22.95 &95.50& -- &95.56 \\
        & \algii & 1.45& -- &20.9 & 94.62& -- &93.69\\
        & \algiuc & 1.42& -- &21.33 &94.66& -- &95.66\\
        & \algic &3.55& -- &21.11 &94.25& -- &95.35\\
        & \algisc & 3.58& -- &20.26&94.25& -- &95.35 \\
        \midrule
        
        % \multicolumn{8}{c}{50-shot} \\
        % \midrule
        % \multirow{5}{*}{20News} & SCL & 40.85 & 49.07 & 59.52 & 58.68 & 66.60 & 74.97 \\
        % & \algii & 32.71 & 41.20 & 56.03 & 58.47 & 65.75 & 74.50 \\
        % & \algiuc & 33.43 & 41.66 & 57.32 & 58.30 & 65.53 & 74 .44\\
        % & \algic & 33.82 & 42.11 & 57.47 & 57.79 & 65.52 & 74.08 \\
        % & \algisc & 33.30 & 40.96 & 56.47 & 57.78 & 65.44 & 74.16 \\
        % \midrule
        % \multirow{5}{*}{WOS} & SCL &0.50& -- &17.15 & 24.23 & -- & 41.31 \\
        % & \algii & 0.50& -- &16.90 & 24.90 & -- & 43.01\\
        % & \algiuc & 21.41& -- &57.08 &52.08& -- &69.80\\
        % & \algic &45.33& -- &61.20 &51.95& -- &69.12\\
        % & \algisc &  \\
        % \midrule
        % \multirow{5}{*}{DBPedia} & SCL &31.03& -- &57.90 &95.50& -- &95.56 \\
        % & \algii & 22.02& -- &51.64 & 94.62& -- &93.69\\
        % & \algiuc & 1.42& -- &21.33 &94.66& -- &95.66\\
        % & \algic &3.55& -- &21.11 &94.25& -- &95.35\\
        % & \algisc & 3.58& -- &20.26&94.25& -- &95.35 \\
        % \midrule

        \multicolumn{8}{c}{100-shot} \\
        \midrule
        \multirow{5}{*}{20News} & SCL & 49.47 & 58.26 & 65.59 & 62.97 & 69.95 & 76.86 \\
        & \algii & 50.70 & 58.22 & 67.07 &63.06&70.42&77.50\\
        & \algiuc & 54.73 & 63.09 & 75.05 &64.23&71.38&78.09 \\
        & \algic & 63.52 & 70.83 & 78.21 &63.21&70.17&76.95\\
        & \algisc & 63.54 & 70.88 & 78.48 &64.49&72.34&78.61 \\
        \midrule
        \multirow{5}{*}{WOS} & SCL &1.17& -- &16.30 &42.65& -- &46.95 \\
        & \algii & 1.19& -- &16.54 & 29.35& -- &46.65\\
        & \algiuc & 37.54& -- &66.61 &51.25& -- &66.97\\
        & \algic &59.59& -- &72.70 &61.14& -- &73.25\\
        & \algisc & 60.02& -- &72.65&62.23& -- &74.56 \\
        \midrule
        \multirow{5}{*}{DBpedia} & SCL &0.06& -- &25.45 &96.03& -- &96.69\\
        & \algii & 1.00& -- &23.72 &96.18& -- &96.83\\
        & \algiuc &84.45& -- &88.10 &95.55& -- &96.69 \\
        & \algic &93.13& -- &94.48 &95.80& -- &96.61 \\
        & \algisc & 93.19& -- &94.63&95.78& -- &96.61 \\
        \bottomrule

        % \multicolumn{8}{c}{300-shot} \\
        % \hline
        % \multirow{5}{*}{20 newsgroups} & SCL & 53.70&59.68&67.41 \\
        % & $\text{LA}_{ii}$ &53.72&61.88&67.96 \\
        % & $\text{LA}_{iuc}$ &66.38&73.10&79.85 \\
        % & $\text{LA}_{ic}$ & 66.80&73.78&80.07\\
        % & $\text{LA}_{isc}$ & 66.62&74.20&80.22 \\
        % \hline
        % \multirow{5}{*}{DBpedia} & SCL  \\
        % & $\text{LA}_{ii}$ &  \\
        % & $\text{LA}_{iuc}$ &  \\
        % & $\text{LA}_{ic}$ &  \\
        % & $\text{LA}_{isc}$ &  \\
        % \hline

    \end{tabular}
    \caption{Results on few-shot in supplement to \Scref{sec:fewshot}.}
    \label{tab:few-shot}
\end{table*}
This section includes the full results in supplement to \Scref{sec:fewshot} shown in Table \ref{tab:few-shot}.

\end{document}